\documentclass[journal]{IEEEtran}

\usepackage{algorithm2e}

\usepackage{amsfonts}
\usepackage[utf8]{inputenc}

\usepackage{footmisc}
\usepackage[noadjust]{cite}

\usepackage{amsmath}
\usepackage{amssymb}
\usepackage{graphicx}
\usepackage{epstopdf}

\usepackage{url}
\usepackage{xcolor}
\usepackage{nicefrac}

\usepackage{hyperref}

\newcommand\nnfootnote[1]{%
  \begin{NoHyper}
  \renewcommand\thefootnote{}\footnote{#1}%
  \addtocounter{footnote}{-1}%
  \end{NoHyper}
}

\newcommand{\etal}{\textit{et al.}~}

\newcommand{\wrt}{with respect to~}
\newcommand{\eg}{e.g.,~}

\newcommand{\x}{\mathbf{x}}

\newcommand{\mcS}{\mathcal{S}}

\newcommand{\mbbR}{\mathbb{R}}

\newcommand{\mcB}{\mathcal{B}}

\newcommand{\mbfz}{\mathbf{z}}
\newcommand{\mbfZ}{\mathbf{Z}}
\newcommand{\mbfx}{\mathbf{x}}
\newcommand{\mbfX}{\mathbf{X}}

\newcommand{\mbfT}{\mathbf{T}}

\newcommand{\mbfu}{\mathbf{u}}
\newcommand{\mbfSigma}{\mathbf{\Sigma}}

\usepackage[noadjust]{cite}

\begin{document}
	
	
	\title{Learning Deep Features for Shape Correspondence with Domain Invariance}
	
	\author{Praful Agrawal, Ross T. Whitaker, Shireen Y. Elhabian}

	\markboth{IEEE Transactions on Medical Imaging}%
	{Learning Deep Features for Shape Correspondence with Domain Invariance}
	
	\maketitle
	
	\nnfootnote{This work is still under review. This work was supported by the National Institutes of Health under grant numbers NIBIB-U24EB029011, NIAMS-R01AR076120, NHLBI-R01HL135568, NIBIB-R01EB016701, and NIGMS-P41GM103545.\\
		Praful Agrawal is an Applied Scientist at Amazon.com. This work was finished during his PhD at the University of Utah, USA. (email: prafulag@cs.utah.edu). Ross T. Whitaker is a Professor and the Director in the School of Computing, University of Utah, Salt Lake City, USA. (email: whitaker@sci.utah.edu). Shireen Y. Elhabian is a Research Assistant Professor in the School of Computing, University of Utah, Salt Lake City, USA. (email: shireen@sci.utah.edu)}

	\begin{abstract}
Correspondence-based shape models are key to various medical imaging applications that rely on statistical analysis of populations of anatomical shapes. Such shape models are expected to represent consistent anatomical features across the population for meaningful, population-specific shape statistics. Early approaches for correspondence placement rely on nearest neighbor search for simpler anatomies. 
Coordinate transformations for shape correspondence hold promise to address the increasing anatomical complexities. Nonetheless, due to the inherent shape-level geometric complexity and population-level shape variation, the coordinate-wise correspondence often does not translate to the anatomical correspondence. 
An alternative, group-wise approach for dense correspondence placement is to explicitly model the 
trade-off between the geometric description of each shape and the statistical compactness of the population. However, due to primitive assumptions, these models achieve limited success in resolving nonlinear shape correspondence. 
Recent works have addressed this limitation by adopting an application-specific notion of correspondence through lifting positional data to a higher dimensional feature space (\eg sulcal depth, brain connectivity, and geodesic distance to anatomical landmarks). 
However, they heavily rely on the manual expertise to create domain-specific features and the consistent placement of landmarks. 
This paper proposes an automated feature learning approach, using \textit{deep} convolutional neural networks, to extract correspondence-friendly features from shape ensembles. 
Further, an unsupervised domain adaptation scheme is introduced to augment the pretrained geometric features with new anatomies.
Results on anatomical datasets of human scapula, femur, and pelvis bones demonstrate that features learned in supervised fashion show improved performance for correspondence estimation compared to the manual features.
Further, the unsupervised learning is demonstrated to learn features for a complex anatomy using the supervised domain adaptation from features learned on a simpler anatomy. 

	\end{abstract}
	
	
	\IEEEpeerreviewmaketitle
	
	\section{Introduction}
\begin{sloppypar}
Statistical shape models have important applications in various
medical imaging tasks, such as image segmentation (\eg
\cite{heimann2009statistical}), hypothesis testing (\eg
\cite{bredbenner2010statistical}), anatomical reconstruction (\eg
\cite{balestra2014articulated}), and pathology detection (\eg
\cite{shen2012detecting}). The utility of such models is influenced
by a shape representation that facilitates
statistical analysis. In this regard, \textit{landmarks} are a 
popular choice as a lightweight but effective representation. Compared to
embedding shapes in the image intensity values at voxels, landmarks-based representation is more intuitive and promotes visual communication
of the results \cite{sarkalkan2014statistical}. To perform shape
statistics, landmarks should be defined consistently within a given
population to refer to the same anatomical position on every shape
instance, a concept known as \textit{correspondence}.  Such
correspondences are often created manually, but this is
time-/labor-intensive, requiring qualified specialists (\eg
radiologists), and cost-prohibitive for 3D images and large
collections of imaging data. 

Advances in shape modeling have achieved some success in automatic placement of correspondences on anatomical structures \cite{cates2017shapeworks,davies2002minimum,durrleman2014morphometry,styner2006framework}, and have been effective in a range of biomedical applications (e.g., \cite{harris2013statistical, atkins2017evaluation, atkins2017quantitative, atkins2019two, bhalodia2019quantifying,kozic2010optimisation,galloway2013large,bryan2009use,zhao2008hippocampus,wang2012comprehensive}). 
However, this success is limited to anatomies that conform to the modeling assumptions in existing methods. Typically, the assumptions include mapping to primitive shapes (such as a sphere) or predefined initial atlas, linear shape variations, and pairwise comparisons with no consideration of population-based shape variations. 
Such assumptions often do not generalize well for complex anatomies \cite{goparaju2018evaluation}. 
The lunate surface of the acetabulum in a pelvis, for example, has a horseshoe-like shape that does not adhere to a standard primitive. 
Pairwise comparisons do not observe the entire population, leading to biased and suboptimal models \cite{oguz2008cortical,oguz2016entropy,balci2007free}, and in many cases fail to find surface-to-surface correspondences across populations.
Anatomical variability can be far more complex than linear approximations, in which case nonlinear variations normally exist, e.g., bending fingers, soft tissue deformations, and vertebrae with different types (lumbar, thoracic, and cervical).
Furthermore, the complexity of a shape anatomy can be quantified as a comparison between the statistical shape variations and the local shape features (such as Feret's diameter \cite{dravzic2016estimation}). 
The complexity of the shape increases as the shape features get smaller compared to the observed variations among different samples. Due to this smaller feature size, the task of establishing correspondence becomes more challenging and thus limits the success of existing methods.

Previous works in the context of brain imaging have included sulcal depth \cite{oguz2008cortical}, brain connectivity \cite{oguz2009cortical}, and anatomical landmarks \cite{datar2013geodesic} as additional domain knowledge to guide the correspondence optimization for complex shapes. 
%
Although promising, these approaches are specific to a particular anatomy and thus not generalizable to other anatomies. Further, their success heavily relies on the domain-specific expertise (and time) to find the geometric features or landmarks consistently on shape samples across the population. Furthermore, such domain knowledge cannot be transferred to new anatomies.

In \cite{agrawal2017learning}, we presented an automated feature-learning to aide the generation of point-wise shape correspondence models. 
The idea is motivated by recent advances in computer vision and computer graphics that use deep convolutional neural networks (CNNs) to learn shape features to establish pairwise correspondences between shapes (e.g., \cite{boscaini2016learning,boscaini2016anisotropic,chopra2005learning}).
In \cite{agrawal2017learning}, shape-specific local geometric features are learned via a surrogate task of correspondence vs noncorrespondence classification for a pair of points sampled from the shape surface(s). At the specific point location, a snapshot of local surface geometry is used as input to the convolutional neural network to facilitate the feature learning. The results indicate that the learned features represent the inherent shape anatomy and are able to successfully guide the optimization of point-wise correspondences. This paper provides an extended description of methods and experimental results presented in \cite{agrawal2017learning}. Further, we propose a methodology to generate geometric features for a new complex anatomy via unsupervised domain adaptation of pretrained features. We use orthopedic data with femur and pelvis shapes to showcase successful feature adaptation.

%
%

\noindent The contributions of this work can be summarized as follows: 
\begin{itemize} 
	\item[--] supervised deep feature learning using \textit{direct surface geometry} rather than relying on predefined surface descriptors;
	

    \item[--] incorporating the \textit{learned features} in the optimization of dense correspondences for complex shape ensembles.
	
	\item[--] unsupervised adaptation of pretrained features to new, complex anatomies; and
	
	\item[--] comprehensive experiments showcasing supervised feature learning and unsupervised feature adaptation.
\end{itemize}

\end{sloppypar}

	\section{Related Work}
\begin{sloppypar}


The relevant work falls into three categories: existing methods used to generate shape correspondence models, deep learning methods used to establish shape correspondence, and domain adaptation schemes, with a particular focus on feature-level adaptation.

\end{sloppypar}

\subsection{Correspondence Estimation}
\label{sec:psm}
\begin{sloppypar}


Correspondence-based shape models represent each surface as a point cloud, where the notion of correspondence is established by the index position of each point. 
In an optimal correspondence model, a particular index should represent the same anatomical location across all the point clouds. The problem of solving point-wise correspondence is challenging due to the variations observed across different shapes. Existing shape modeling approaches to estimating correspondences fall into two broad families: a \textit{pairwise} approach that relies on pairwise shape comparisons and a \textit{groupwise} approach that observe the entire population. 

Pairwise methods that rely on finding nearest points between surfaces (e.g. \cite{icp} and their variants) work well for limited anatomies with large sized shape features but are prone to mismatches with complex geometries having high curvature regions and small sized shape features. 
%
Alternatively, other pairwise methods try to match predefined parameterizations of surfaces while maintaining regularity.  
For example, the spherical harmonics point distribution model (SPHARM-PDM) \cite{styner2006framework} relies on a smooth
one-to-one mapping from each shape instance to the unit sphere. The mapping to a predefined surface topology and the a priori assumptions on smoothness, rather than population-specific shape features, limit the class of anatomies and populations for shape modeling. 
%
Correspondence estimation based on pairwise coordinate transformations (e.g., \cite{durrleman2014morphometry,joshi2004unbiased,joshi2000landmark}) holds promise, but they fail to adapt the deformation metric to the inherent modes of variability in the population, and in many cases fail to find surface-to-surface correspondences across entire populations.

Groupwise approaches use the statistics of the population itself to drive the matching of shape features.  
For instance, minimum description length (MDL) \cite{davies2002minimum} optimizes point correspondences using an information content objective, but it relies on intermediate spherical surface parameterizations,  which places limitations on the types of shapes and the optimization process. 
Forgoing the parameterization, the particle-based shape modeling (PSM) approach has offered a flexible nonparametric and general framework to establish dense point correspondences across shape ensembles without constraining surface topology or
parameterization \cite{cates2007shape,oguz2016entropy,cates2017shapeworks}. The PSM approach explicitly models the inherent trade-off between the geometric description of each shape and the statistical compactness of the population via an entropy optimization scheme.
Recently, it has been show that the PSM approach outperforms other pairwise \cite{styner2006framework} and groupwise \cite{durrleman2014morphometry} methods in consistently estimating anatomical measurements and capturing clinically relevant population-level shape variations \cite{goparaju2018evaluation}.
In this paper, we thus focus on PSM and its variants to demonstrate the efficacy of learned shape features for correspondence optimization.

Current PSM implementations rely on a Gaussian model in the {\em shape space}, which is the vector space formed
by the spatial coordinates of surface correspondences (modulo a similarity transform).
However, the distribution of anatomical structures can be far more complex than the Gaussian
assumes, and surface locations are not always indicative of their correspondence. To address these shortcomings, Datar \etal \cite{datar2013geodesic} proposed to use geodesic distances to user-identified landmarks on individual shapes. These point-wise distance values were used to guide the entropy-based optimization of shape correspondence. Further, Oguz \etal (in the context of brain imaging) considered sulcal depth \cite{oguz2008cortical} and brain connectivity \cite{oguz2009cortical} as additional features in the PSM optimization. While promising, such approaches  are tailored to a particular dataset, anatomy and application. 
%
In \cite{agrawal2017learning}, we proposed to automatically learn the shape-specific anatomical features in supervised fashion. Here, we propose to adapt the pretrained features for new anatomies.

\end{sloppypar}

\subsection{Deep Learning for Shape Correspondence}
\begin{sloppypar}


Shape matching methods in computer vision and computer graphics represent shapes as 3D meshes, and use point-wise matching to establish correspondence between a pair of shapes. Here, we focus on scenarios where point-wise matching is simultaneously optimized with population-level statistics.

Recent works have trained deep convolutional neural networks (CNNs) to estimate the point-wise matching between an input 3D mesh and a template mesh \cite{van2011survey}. 
Boscaini \etal \cite{boscaini2016anisotropic} proposed a supervised learning of pointwise shape correspondence using anisotropic CNNs. The network was trained using a predefined correspondence model. Thus, this method requires retraining for a new set of correspondences. 
Groueix \etal \cite{groueix20183d} estimated shape correspondence with respect to a predefined template using feature descriptors. 
The choice of template shape impacts the resulting statistical variations, and thereby does not lead to a generic population-based statistical model. 
Kleiman and Ovsjanikov \cite{kleiman2019robust} adopted a graph matching approach to establish region-wise correspondence across shapes. The method works with any given shape representation and relies on inherent symmetric nature of shapes. This symmetric property limits the success for shapes with significantly varying topologies. Sun \etal \cite{sun2017deep} trained a Siamese network for a correspondence vs noncorrespondence classification of an input pair. The training relies on predefined spectral shape descriptors to enable shape matching. Although predefined descriptors may not generalize to complex shapes, this method does not require selection of a template. Hence, we adopt this idea of a Siamese network for a correspondence vs noncorrespondence classification. As input, we have chosen to use a direct snapshot of local surface geometry in place of spectral descriptors, enforcing the network to automatically learn the shape features.
\end{sloppypar}

\subsection{Domain Adaptation}
\begin{sloppypar}
Here, our goal is to learn a set of geometric features that may generalize to unknown classes of shapes. The supervised learning of such generic features requires a large pool of exhaustive training data. 
Further, for complex shapes, the labeled training pairs are often expensive and require manual expertise. 
Thus, to alleviate the need for labeled data, we refine the pretrained features on simple anatomies to model new anatomies in an unsupervised way borrowing ideas from unsupervised domain adaptation techniques.
%

A recent survey on domain adaptation \cite{wang2018deep} illustrates the different techniques to learn semantically meaningful and domain invariant features. Broadly, the domain adaptation techniques follow either adversarial training paradigms or a feature space matching approach. In the feature space matching, the source and target distributions are made to align in the output feature space. For instance, Sun \etal \cite{sun2016return} formulated the problem as minimizing the distance between covariance matrices of source and target domain feature spaces, thus resulting in aligning the source domain feature space with that of the target domain. Fernando \etal \cite{fernando2013unsupervised} used subspace matching to map the input data from the source domain onto the target domain. Although simple, Fernando \etal's approach is dependent on the correct subspace estimation, which typically requires a large number of samples. In the case of a limited number of samples in either or both domains, these methods may lead to overfitting.

Adversarial training is a more generic approach where a discriminator function is trained on the output features, with an adverse objective function. The goal of this adversarial objective is to promote confusion for the discriminator function, to distinguish between source and target domains. There are two types of models for adversarial training; 1) based on a generative model, which tries to generate impostor samples for the discriminator network \cite{bousmalis2017unsupervised, isola2017image, liu2016coupled}; and 2) based on learning discriminative features by directly using them as inputs to the discriminator network. The goal is to ensure that the output features perform poorly at discriminating between samples from the two domains, while being good at the original task \cite{ganin2014unsupervised, ganin2016domain, tzeng2017adversarial, tzeng2015simultaneous, tzeng2015adapting}.

In this paper, we adopt the domain adversarial training proposed by Ganin and Lempitsky \cite{ganin2014unsupervised}, where the output features are used to solve a parallel domain classification problem. The sign-reversed gradients from the augmented discriminator network are backpropagated to encourage domain confusion.
\end{sloppypar}
%
%
%
%
%

	\section{Methodology}

\begin{sloppypar}


In this paper, we propose to train a deep neural network to learn features that capture the geometry of any given anatomy. Specifically, the features should highlight the shape properties of the anatomy, such as convexity/concavity, saddle structures, and flat regions. Such a feature bank can help correlate anatomical regions across different shape samples, and thus guide the correspondence optimization. We devise a supervised feature learning scheme via a paired-classification task. As input to the deep neural network, we extract local surface geometrical characteristics using circular geodesic patches in the neighborhood of given point locations on a shape's surface (see Figure \ref{fig:patchIllustration}). Once trained, the network should encode similar anatomical regions with similar feature vectors. We further use the unsupervised domain adversarial training \cite{ganin2014unsupervised} to adapt the trained network for new anatomies. The trained feature maps, supervised and/or unsupervised, are then used to aide shape correspondence optimization in the context of particle-based shape modeling (PSM) \cite{cates2007shape,cates2017shapeworks}.

In this section, we discuss the details of
input patch extraction, followed by the deep network configurations used for supervised feature learning and unsupervised feature adaptation to new anatomies. We conclude the section
with details to incorporate the extracted features into an existing method for correspondence optimization and evaluation criterion used for the resulting shape-correspondence models.
\end{sloppypar}
%

%
%


\subsection{PSM Background}
Particle-based shape modeling (PSM) is a method for constructing compact statistical point-based models of shape ensembles while not relying on any specific surface parameterization \cite{cates2006entropy,cates2007shape,cates2008particle,oguz2016entropy}. PSM uses a set of dynamic particle systems, one for each shape, in which particles interact with one another with mutually repelling forces to cover optimally, and therefore describe the surface geometry. An image segmentation process is generally used to extract the anatomy of interest in which the surface geometry is described implicitly as the interface between foreground and background regions. PSM makes use of the signed distance transform (DT) of this binary representation for finite numerical calculation of surface geometry. Here we give a brief review of the PSM method.

\vspace{0.02in}
\noindent\textbf{PSM formulation:} Consider a cohort of shapes $\mcS = \{\mbfz_1, \mbfz_2, ..., \mbfz_N\}$ of $N$ surfaces, each with its own set of $M$ corresponding particles $\mbfz_n = [\mbfz_n^1, \mbfz_n^2, ..., \mbfz_n^M] \in \mbbR^{dM}$ where each particle $\mbfz_n^m \in \mbbR^d$ whose ordering implies correspondence among shapes (typically $d=3$ for anatomical structures). This representation involves \textit{two types of random variables}: a shape space variable $\mbfZ \in \mbbR^{dM}$ and a particle position variable $\mbfX_n \in \mbbR^d$ that encodes particles distribution on the $n-$th shape (configuration space). For groupwise modeling, shapes in $\mcS$ should share the same world coordinate system. Hence, generalized Procrustes alignment \cite{gower1975generalized} is used to estimate a rigid transformation matrix $\mbfT_n$ that can transform the particles in the $n-$th shape local coordinate $\mbfx_n^m$ to the world common coordinate $\mbfz_n^m$ such that $\mbfz_n^m = \mbfT_n\mbfx_n^m$. Correspondences are established by minimizing a combined  shape correspondence and surface sampling cost function $Q = H(\mbfZ) - \sum_{n=1}^N H(\mbfX_n)$,  where $H$ is an entropy estimation assuming Gaussian shape distribution with covariance $\mbfSigma$ in the shape space and Euclidean particle-to-particle repulsion in the configuration space \cite{cates2007shape}. In particular, $H(\mbfZ) \approx \nicefrac{1}{2} \log |\mbfSigma| = \nicefrac{1}{2} \sum_{j=1}^{N} \log \lambda_j$ where $\lambda_j$ are the eigenvalues of $\mbfSigma$. Since $N \ll dM$, covariance is estimated in the dual space of dimension $N$ which defines the hyperplane in $\mbbR^{dM}$ where all the $N-$samples inhabit. This formulation favors a compact ensemble representation in shape space (first term) against a uniform distribution of particles on each surface for accurate shape representation (second term). The optimization process is defined via gradient descent, by moving individual points on the surface to minimize $Q$, as described in \cite{cates2007shape}.

\subsection{Geodesic Patch Extraction}
\begin{figure}[!b]
	\centering
	\includegraphics[width=0.95\columnwidth]{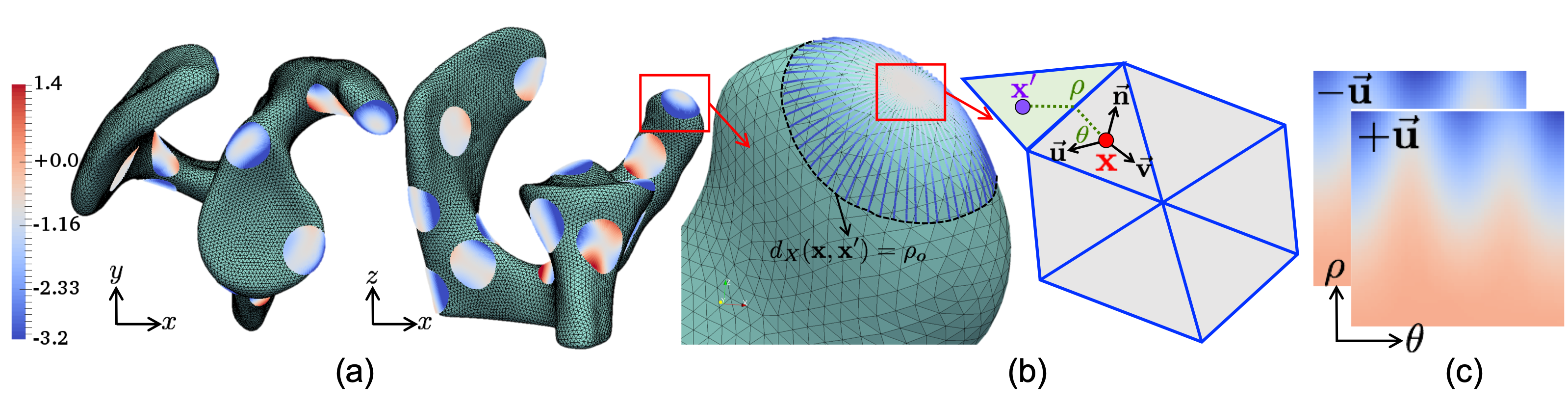}
	\caption{Geodesic patch extraction at a surface point $\mbfx$:
		(a) Sample patches on a human scapula mesh. (b) Finding
		a point ($\mbfx'$ at a geodesic distance $\rho$ in
		direction that makes angle $\theta$ with first principal
		curvature direction ($\mbfu$) in the tangent space of
		$\mbfx$. (c) Two channels of an input to the CNN,
		representing geodesic patches where every pixel corresponds
		to the signed normal distance of point $\mbfx'$ in the patch to the tangent plane at $\mbfx$.} 
	\label{fig:patchIllustration}
\end{figure}
\begin{sloppypar}
The success of deep convolutional neural networks (CNNs) has been demonstrated in analyzing functions defined on Euclidean
grid-like domains such as images. Nonetheless, non-Euclidean data,
particularly surface geometry, do not directly benefit from these
deep models in which operations such as convolution and pooling are
not readily defined. However, using a 3D patch from a volumetric shape representation (\eg label maps or signed distance maps) to represent the local surface geometry may not yield optimal results. Furthermore, using 3D convolutions will result in a very large number of network parameters with a limited field of view.

Recently, local charting methods (\eg
\cite{boscaini2016learning}) have been proposed as a generalization of
convolution to non-Euclidean domains where a \textit{patch operator}
is defined to extract the local surface patches that are subsequently
associated with predefined \textit{shape descriptors}. Here, we rely
directly on surface geometry to compute such local shape
descriptors. Specifically, a spatial function of surface distance to the tangent space can
encode local geometrical properties in a local neighborhood. Hence, we propose
to use signed normal distance to the tangent space sampled in the
geodesic neighborhood of a surface point $\mbfx$ as a snapshot
of the local surface geometry. We use the principal curvature directions $(\vec{u}, \vec{v})$ to define
the local intrinsic coordinate system at the
surface point $\mbfx$. As illustrated in
Figure~\ref{fig:patchIllustration}, neighboring points that lie in the
geodesic ball $\mcB(\mbfx) = \{\mbfx' : d_X(\mbfx, \mbfx') \leq
\rho_o\} $ with a radius $\rho_o > 0$ are sampled on a circular geodesic
patch. Each ring in the patch corresponds to a particular geodesic distance
$d_X(\mbfx, \mbfx') = \rho$ with $\rho \in [0, \rho_o]$. Total 64 concentric rings are sampled with  radii increasing by a fixed geodesic distance upto a maximum of $rho_o$. Every ray
originating from $\mbfx$, perpendicular to the geodesic rings, is
inclined at an angle $\theta \in [0,2\pi)$ to the first principal
curvature direction at $\mbfx$. Total 64 rays are uniformly sampled to cover the $2\pi$ angle. Thus, resulting in a rectangular patch of size $64\times64$. The parameter $\rho_o$ is set to $5\%$ of the maximum shape diameter in the ensemble.

Principal curvature
directions are estimated using least squares fitting of a quadratic
\cite{rusinkiewicz2004estimating}. We enforce the right-hand rule on
local coordinates with the normal direction representing the z-axis. 
However, the principal curvature
directions are accurate only up to the sign of the vector. To address this ambiguity, we
extract two patches per surface point (see
Figure~\ref{fig:patchIllustration}c), one with bases defined by the positive sign of principal directions and the other by negative sign, and use them as a two-channel image for input to the neural network. The other two combinations with opposite signs of the two directions are invalid as they violate the right hand rule.

There still remains ambiguity about order of the two patches in the two-channel input image. We perform the supervised training of paired-classification such that the network is invariant to this ambiguity. Thus we use all four possible combinations of an input pair as independent training samples to help the network become invariant to this sign ambiguity.
\end{sloppypar}

\subsection{Supervised Feature Learning}
\begin{sloppypar}
We use deep CNNs to
learn local shape features for establishing shape correspondence. The
goal is to make some of the hidden layers of the network respond
similarly at corresponding locations across the given shape ensemble,
and dissimilarly at noncorresponding ones. Therefore, we design the feature learning problem
as a paired-classification task (corresponding vs noncorresponding).

\begin{figure}[!t]
	\centering
	\includegraphics[width=\columnwidth]{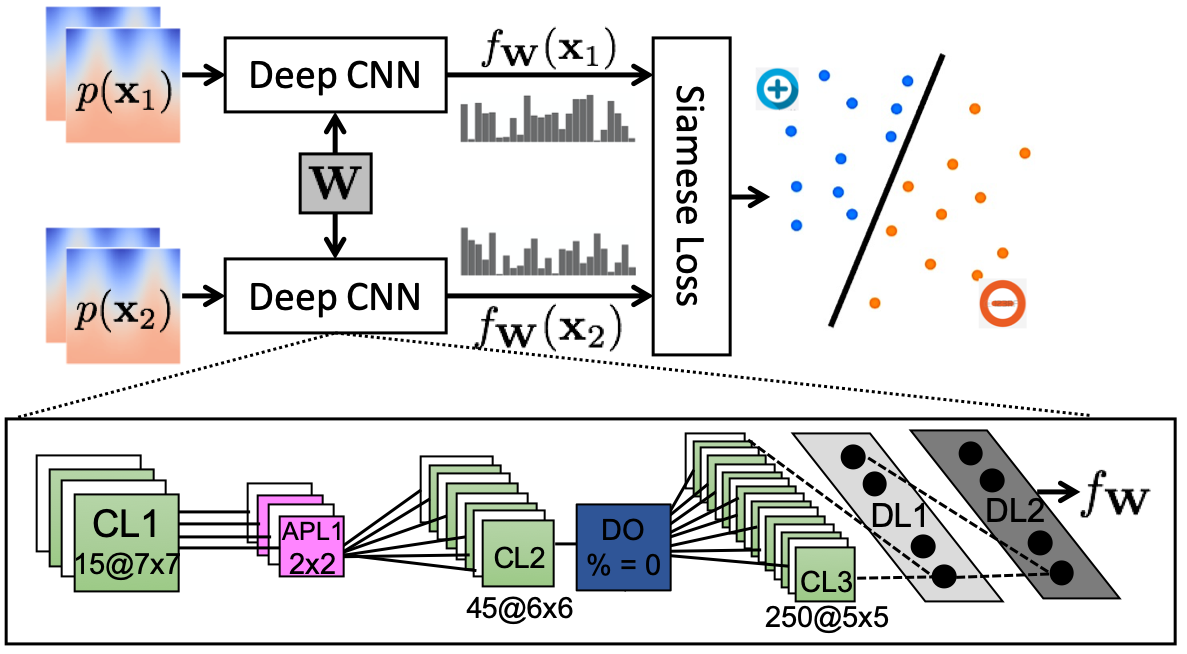}
	\caption{The Siamese network architecture used to learn features for correspondence.}
	\label{fig:siameseCNN}
\end{figure}

Given a pair of inputs, the task is to identify if they belong to a 
similar anatomical location. To accomplish this task, we use a
Siamese network configuration, which
consists of two identical copies of the same deep CNN. 
We train the Siamese CNN
to learn the binary classification problem, where the positive class indicates
correspondence between a pair of input patches, and the negative class suggests
that patches possess different shape characteristics. Contrastive loss \cite{chopra2005learning} is optimized to facilitate the Siamese training. 
After training, one copy of the two identical deep CNNs in the Siamese network becomes a nonlinear filter bank (Figure \ref{fig:siameseCNN}) that is computed on all mesh vertices to produce feature maps on the surface.
We rely on a given point-cloud based correspondence model to generate the supervised training samples, i.e., the corresponding and noncorresponding pairs. The training data consists of a small set of quality controlled point clouds generated using position-based \cite{cates2006entropy} and geodesics-based \cite{datar2013geodesic} PSM methods. The geodesic patches extracted at the point locations are fed as inputs to the Siamese network.

For the CNN configuration, we modify the
network configuration of Chopra \etal \cite{chopra2005learning} with added regularization using batch normalization and dropout layers; see Figure~\ref{fig:siameseCNN}. Beyond the regularization layers, the adapted network configuration includes following updates:
\begin{itemize}
    \item In our experiments, we found that using dropout layer does not impact the classification performance; therefore, we set the probability of dropout to zero.
    \item Batch normalization is used at the end of convolutional layers.
    \item It is critical for the correspondence optimization methods that the output features exhibit smooth transitions from one shape feature to another. Therefore, we replace the max pooling layers with the average pooling.
    \item We use the softplus activation function in place of ReLU. Softplus activation ensures smooth output features, while maintaining faster convergence behavior similar to ReLU.
    \item All the weights are initialized using a zero mean normal distribution with $0.05$ standard deviation, and bias is initialized with zero.
    \item We subtract the mean of the training data from all training patches (and the same mean is subtracted from testing patches for feature extraction) such that the input data can align with the distribution of initial network weights (spanning the hypersphere).
\end{itemize}
\end{sloppypar}

\subsection{Unsupervised Feature Adaptation}
\begin{sloppypar}

The supervised feature learning scheme relies on a given correspondence model to provide training correspondences for the Siamese network. Positional PSM or any other shape modeling approach can provide such training samples. Nonetheless, for highly variable anatomical populations, optimal correspondences are not expected, hence the purpose of learning deep features. In this regard and to alleviate the need for manual intervention through the selection of anatomical landmarks for network training, we make use of recent advances in machine learning for domain adaptation, in which correspondence models optimized for simple varying anatomies (or even synthetic shape ensembles with analytical correspondences), a.k.a. source domain, will leverage populations of unknown correspondences, a.k.a. target domain, with deep learned features that are invariant w.r.t. the shift in shape distribution between the two domains.

\begin{figure}[!t]
	\centering
	\includegraphics[width=0.95\columnwidth]{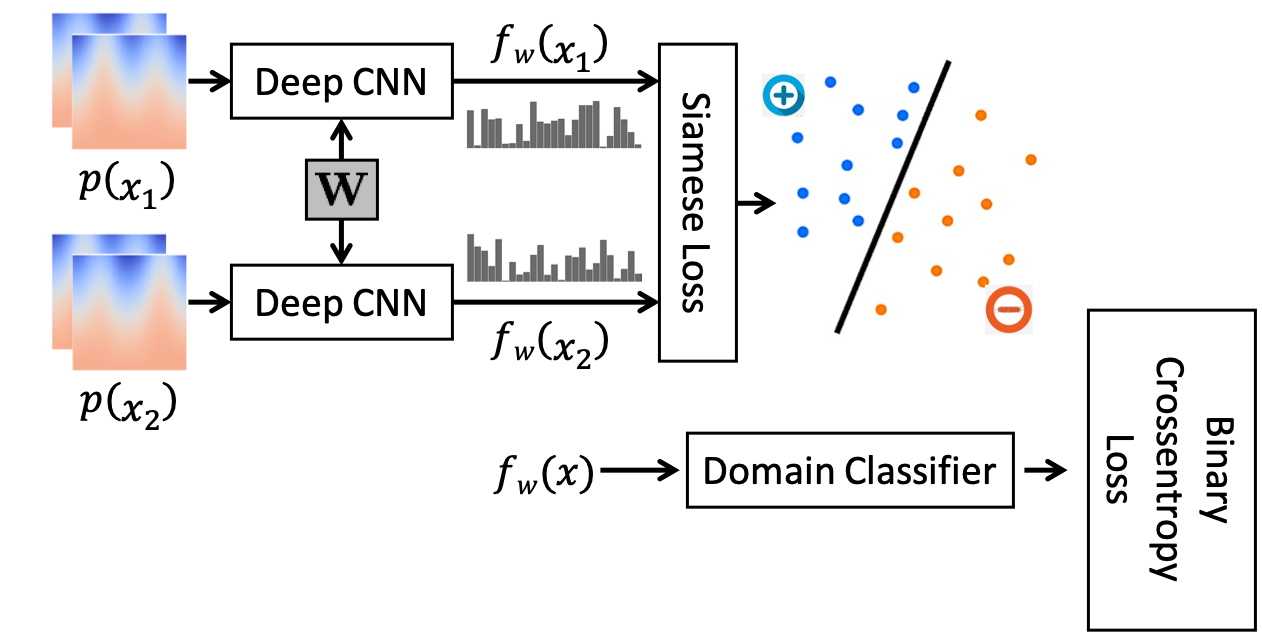}
	\caption{The combined network configuration used for adversarial training of the Siamese network.}
		\label{fig:myDann}
\end{figure}

Given a pretrained feature (Siamese) network for a particular anatomy $S$, the objective is to generalize the output features to a new anatomy, without providing the labeled corresponding/noncorresponding samples for this new anatomy $T$. To accomplish this objective, we adopt the idea of adversarial training, proposed by Ganin and Lempitsky \cite{ganin2014unsupervised}. In \cite{ganin2014unsupervised}, feature-level domain adaptation is achieved by augmenting a domain classifier onto the original learning network (Siamese network in our case, see Figure~\ref{fig:myDann}). The domain classifier is tasked to identify the domain of an incoming sample, represented by the feature vector $f_w(\x)$. The gradients from this domain classifier network are backpropagated into the Siamese network after a sign reversal, typically performed by a gradient reversal layer. This sign reversal facilitates adversarial training of the domain classifier network, and makes it difficult to identify the domain of the input sample. As a result of this confusion, the CNN in the Siamese network is forced to update the parameters so as to combine the topology of the new anatomy $T$ with the shape characteristics of $S$ that is being learned via supervision.

Similar to the configuration used by Ganin and Lempitsky \cite{ganin2014unsupervised}, we use a two-layered network for the domain classification. The two layers contain 500 and 1000 convolution filters of size $1 \times 1$. The binary crossentropy loss function is used for the domain classifier ($Loss_{domain}$) with a weight $\lambda$. Combined with the contrastive loss function \cite{chopra2005learning} as $Loss_{siamese}$, the total loss for the joint network becomes, $Loss_{total} = Loss_{siamese} + \lambda Loss_{domain}$. The value of $\lambda = 0.01$ achieved stable training of the combined network. The Siamese part of the network is initialized using the weights from pretrained network. Glorot initialization \cite{glorot2010understanding} is used for the domain classifier. The initialized network is trained to minimize the total loss, using the samples from $S$ and $T$. The samples from $S$ have corresponding/noncorresponding labels as well as the domain identifying label (i.e., source domain), whereas samples from $T$ possess only domain identification (i.e., target domain). At the end of training, the CNN is expected to produce features that capture shape characteristics from both $S$ and $T$ anatomies.
\end{sloppypar}

\subsection{Deep Feature PSM}
\begin{sloppypar}
We incorporate our learned features into the framework of entropy-based PSM \cite{cates2007shape,cates2017shapeworks}. The method uses a set of dynamic
particle systems, one for each shape, in which particles interact with
one another using mutually repelling forces to sample (and
therefore describe) the surface geometry. Consider a cohort of shapes
$\mcS = \{\mbfz_1, \mbfz_2, ..., \mbfz_N\}$ containing $N$ surfaces,
each with its own set of $M$ corresponding particles. Each shape vector $\mbfz_n =
[\mbfz_n^1, \mbfz_n^2, ..., \mbfz_n^M] \in \mbbR^{dM}$ such that
$\mbfz_n^m \in \mbbR^d$ is the position vector of $m-$th particle on $n-$th shape, and ordering implies correspondence across
shapes. 
Correspondences are established by
minimizing an entropy-based cost function that consists of shape correspondence and surface sampling cost $Q = H(\mbfZ) - \sum_{n=1}^N H(\mbfX_n)$. 
Here, $H(\mbfZ)$ measures the
differential entropy, assuming multivariate normal distribution over samples in the $dM$-dimensional shape space. 
Minimizing this entropy enables correspondence among shape vectors.
Further, $H(\mbfX_n)$ is a nonparametric density estimator in the
(3D) configuration space of the $n-$th shape. Maximizing this entropy ensures uniform sampling of the $n-$th shape surface.
For groupwise modeling, shapes in $\mcS$
should share the same world coordinate system. Generalized
Procrustes alignment \cite{gower1975generalized} is used to estimate a rigid
transformation matrix $\mbfT_n$ per shape instance, such that
$\mbfz_n^m = \mbfT_n\mbfx_n^m$. 
In particular,
correspondence entropy relies solely on particle positions to achieve
compact ensemble representation in shape 
space (first term) against a
uniform distribution of particles on each surface for accurate shape 
representation (second term).




In this paper, we propose to modify the correspondence term $H(\mbfZ)$ by incorporating the 
 features extracted from a trained CNN. Optimizing the updated objective function results in a compact statistical shape model for complex anatomies.
 The extracted features
encode only local geometric properties and require position
information, unlike geodesic distances used in \cite{datar2013geodesic}, which can encode global shape
characteristics. Therefore, we also keep the particle positions to guide optimization for locating corresponding points.
The updated shape vector becomes $\mbfz_n = [\mbfz_n^1, ..., \mbfz_n^M]^T
\in \mbbR^{(d+L)M}$, where $\mbfz_n^m
= \left[ f^1(\mbfx_n^m), ..., f^L(\mbfx_n^m),
(\mbfT_n \mbfx_n^m)^T \right], $ and $f^l(\mbfx_n^m)$ is the
$l-$th deep feature extracted at the $m-$th particle on the $n-$th
shape. 
\end{sloppypar}

\subsection{Evaluation Criterion}
\begin{sloppypar}
We assess the optimized correspondence models using qualitative inspection and quantitative metrics. The qualitative assessment of shape correspondence models entails examining the anatomical correctness and integrity of the mean shape. Further, modes of variation, computed using principal component analysis (PCA), should reveal population-specific variations. 
%
For a quantitative assessment, the resulting shape generative model is evaluated. To construct the generative model, a low-dimensional PCA subspace is learned from the high-dimensional shape space. This generative model is evaluated using the following quantitative metrics:
\begin{itemize}
	\item \textit{Variance plot: }We plot the percentage of variance \wrt PCA (principal component analysis) modes to demonstrate the compactness of sample distribution in the resulting subspace (higher is better).
	
	\item \textit{Generalization: }Proposed by Davies \etal \cite{davies2008statistical}, this quantifies the ability of the generative model to represent the unknown samples. The reconstruction error of the correspondence model for an unseen shape instance, using the PCA model built from training samples, is evaluated in a leave-one-out fashion. We plot the mean of the point-wise physical distance between test samples and their reconstruction \wrt the number of PCA modes (lower is better).
	
	\item \textit{Specificity: }Proposed by Davies \etal \cite{davies2008statistical}, this measures the ability of the generative model to produce plausible shapes. It is quantified as the Euclidean distance between a sampled shape (from PCA model built using all training samples) and the closest training sample (lower is better). The average over 1000 samples is used.
\end{itemize}
\end{sloppypar}
	\section{Results and Discussion}
\begin{sloppypar}
We showcase the results of supervised feature learning and unsupervised adaptation of pretrained features using synthetic and clinical datasets. First, we demonstrate the success of supervised-learned features in 
establishing a good shape-correspondence model for complex anatomies. We then 
showcase the adaptation of features learned from a simpler anatomy to a complex anatomy.
\end{sloppypar}
\subsection{Shape Correspondence Models Using Supervised Features}
\label{sec:results1}
\begin{sloppypar}
\begin{figure}[!ht]
	\centering
	\includegraphics[width=0.85\columnwidth]{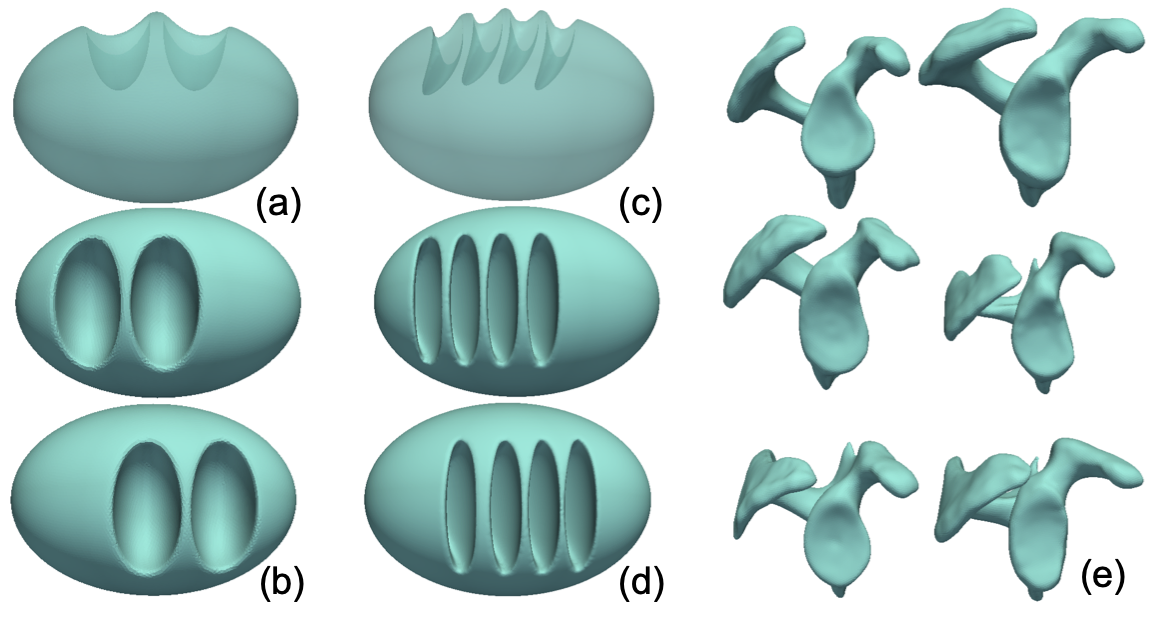}
	\caption{Sample shapes from synthetic and clinical datasets: (a) side view and (b) top view of Bean2, (c) side view and (d) top view of Bean4, (e) scapula. Left column -- controls, right column -- patients.}
		\label{fig:DataSamples}
\end{figure}

Two synthetic datasets and a clincal dataset are used to demonstrate the ability of deep feature PSM to generate improved 
statistical models for complex shapes compared to relying on positional or handcrafted features. The proposed method (deep feature PSM) is compared with positional PSM \cite{cates2007shape,cates2017shapeworks} and geodesic PSM \cite{datar2013geodesic}.

\begin{figure}[!t]
	\centering
	\includegraphics[width=0.8\columnwidth]{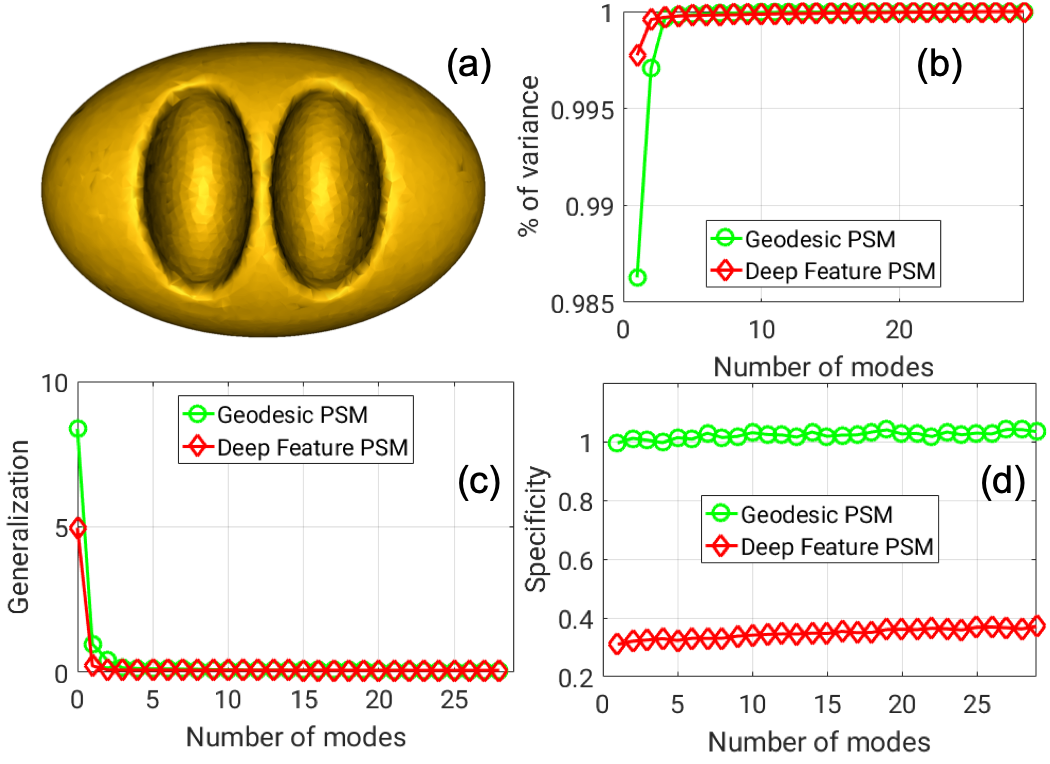}
	\caption{Bean2 results: (a) mean shape from deep feature PSM, (b) variance plot, (c) generalization (mm), (d) specificity (mm), voxel size=1mm.}
	\label{fig:bean2Results}
\end{figure}
\begin{figure}[!t]
	\centering
	\includegraphics[width=\columnwidth]{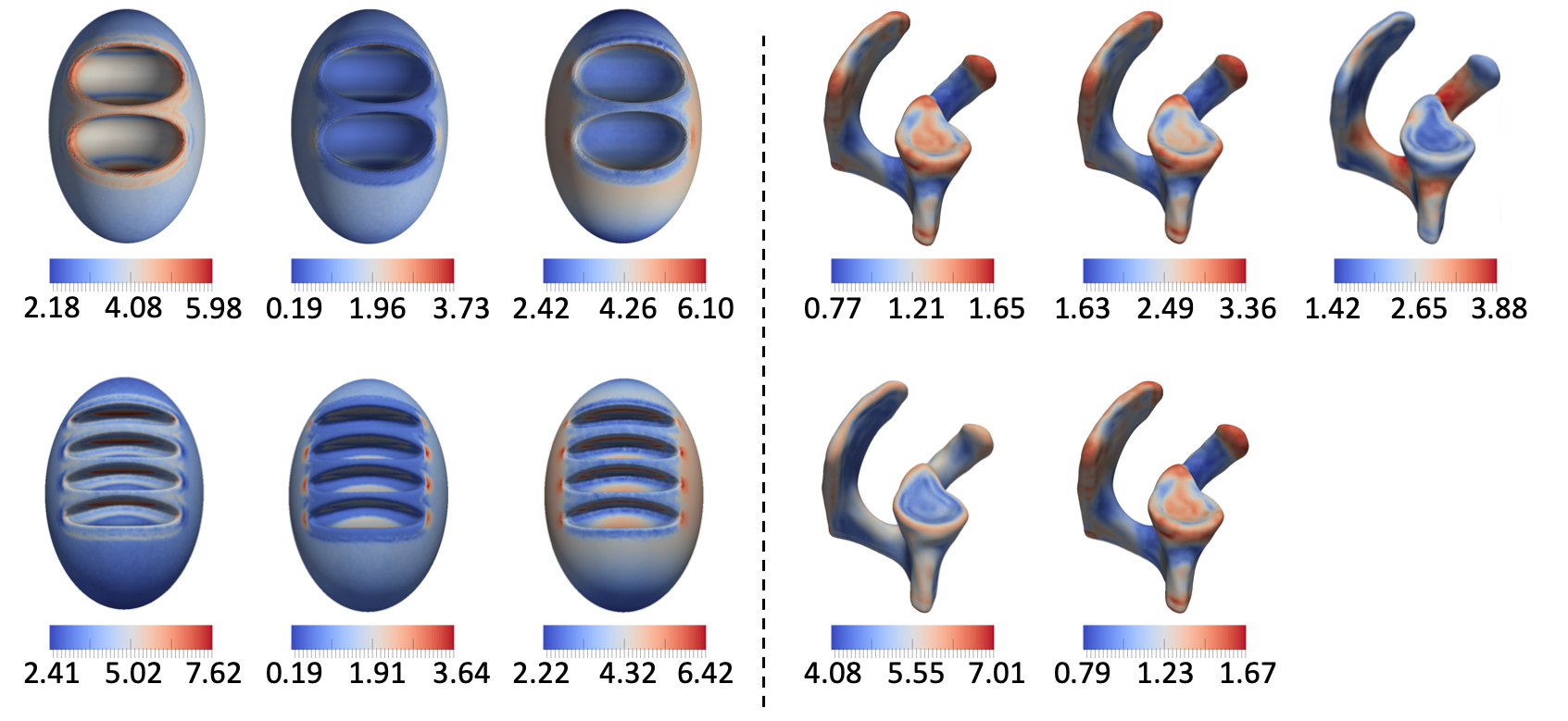}
	\caption{Sample feature maps from networks trained using Bean2 and Scapula datasets.}
	\label{fig:cnnFeatures}
\end{figure}
\begin{figure}[!h]
	\centering
	\includegraphics[width=0.8\columnwidth]{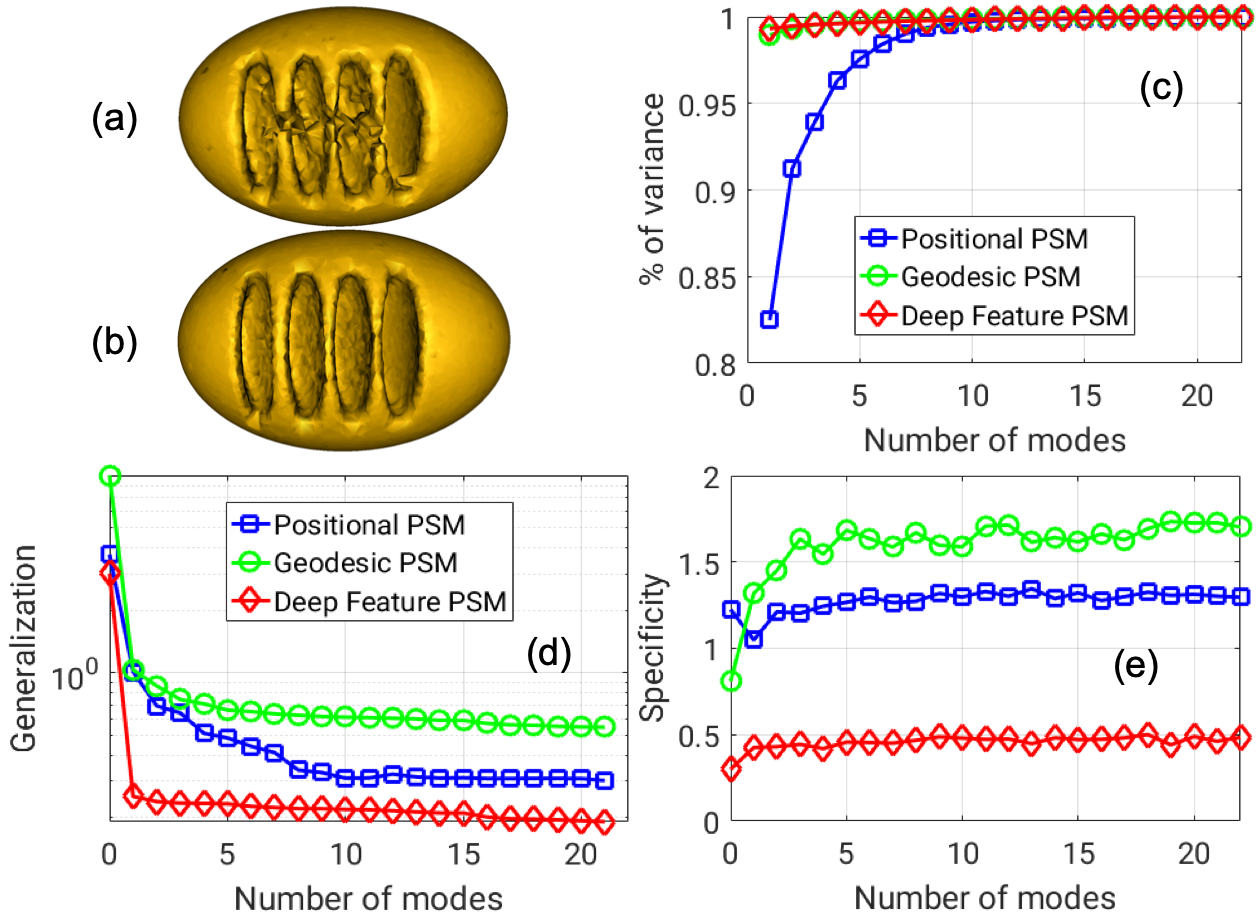}
	\caption{Bean4 results: (a) mean shape from positional PSM, (b) mean shape from deep feature PSM, (c) variance plot, (d) generalization (mm), (e) specificity  (mm), voxel size = 1 mm.}
	\label{fig:bean4Results}
\end{figure}

\subsubsection{Proof-of-Concept} 
The first synthetic dataset (Bean2) contains $30$ shapes that represent a coffee bean with a spatially varying structure. The second dataset (Bean4) is comprised of $23$ samples of a more complex coffee bean shape, with a group of closely located thin structures collectively varying in position (see Figure~\ref{fig:DataSamples}). To generate training samples for the Siamese network, we use an optimized statistical shape model with $3072$ points for Bean2, obtained using the geodesic PSM \cite{datar2013geodesic}. Patches extracted using correspondence points from 6 randomly selected shapes are used to generate the training data. The CNN configuration with $L=10$ output features yields an optimal Siamese classification performance of $0.92$ AUC (area under the curve) of the ROC (receiver operating characteristics) curve, resulting in a $90\%$ true positive ratio (TPR) at the expense of a $20\%$ false positive ratio (FPR). Figure~\ref{fig:cnnFeatures} shows sample features. Given that there are multiple regions with similar shape characteristics, which may lead to a higher FPR, $20\%$ is a relatively small penalty. Moreover, using position information in deep feature PSM will help reduce the impact of false positives. All $10$ features are used in PSM to generate a shape model with $4096$ correspondence points. Figure~\ref{fig:bean2Results} presents the mean shape and quantitative evaluation of the correspondence model. Results indicate compactness of the statistical shape model and better generalization and specificity performance over the geodesic PSM. 

Bean4 shapes have similar characteristics to those of Bean2, and therefore, we use the same trained Bean2 network to extract features for Bean4 shapes (i.e., without feature adaptation). Figure~\ref{fig:cnnFeatures} presents samples of the same feature on the two datasets. Comparative results on Bean4 data, presented in Figure~\ref{fig:bean4Results}, highlight the better performance of the proposed method over positional PSM in generating a more compact statistical shape model. The generative model from the proposed method also outperforms in its ability to generalize over unseen data and to produce plausible shape samples. 

\begin{figure}[!b]
	\centering
	\includegraphics[width=0.8\columnwidth]{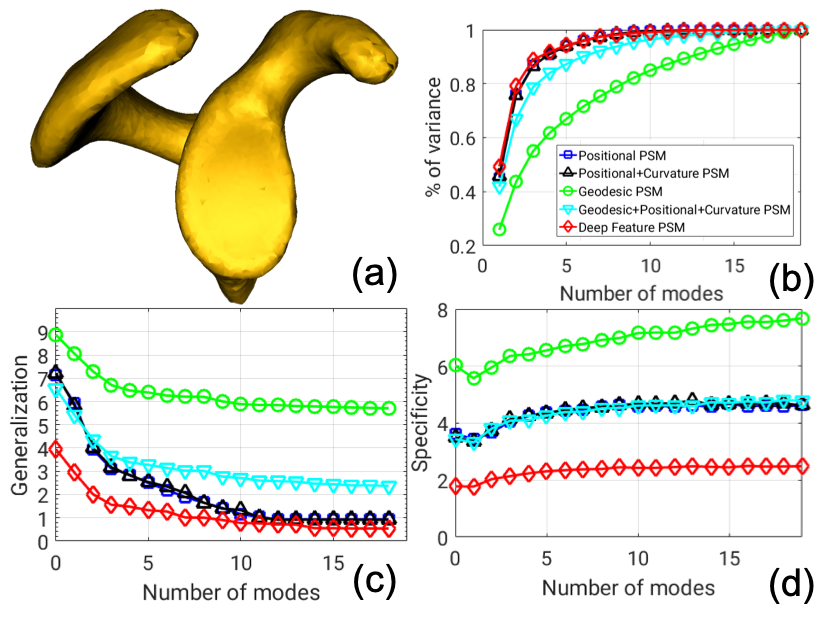}
	\caption{Scapula results: (a) mean shape from deep feature PSM, (b) variance plot, (c) generalization (mm), (d) specificity (mm), voxel size = 0.5 mm.}
		\label{fig:scapulaResults}
\end{figure}

\subsubsection{Clinical Data} 
A clinical dataset of $20$ scapula shapes ($10$ controls and $10$ patients with osseous Hill-Sachs lesions) is used for experiments. Samples are rigidly aligned \wrt the glenoidal plane, and a set of $16$ anatomical landmarks are defined manually. Reconstructed scapula meshes are then clipped to the glenoid, acromion, and coracoid to model the areas of high geometric curvature related to constraint of the humeral head; Figure~\ref{fig:DataSamples} illustrates sample shapes. The shape model with $2432$ points and 6 randomly selected shapes are optimized using the geodesic PSM to generate the training data. Using $L=5$ output features gives optimal classification performance of AUC$ = 0.80$. Figure~\ref{fig:cnnFeatures} shows sample feature maps.

For additional comparison, we augment geodesic and positional PSM with local surface curvature.
Figure~\ref{fig:scapulaResults} showcases the mean shape from correspondence model generated using the proposed deep features, and shows the improved quantitative measures in comparison to handcrafted features. It is important to note that the proposed method is able to achieve generalization and specificity of about $2$ voxels using dominant modes ($5$ modes) in contrast to a minimum of $3$ voxels from other methods.
\end{sloppypar}

\subsection{Shape Correspondence Models Using Adapted Features}
\begin{sloppypar}
\begin{figure}[!ht]
	\centering
	\includegraphics[width=\columnwidth]{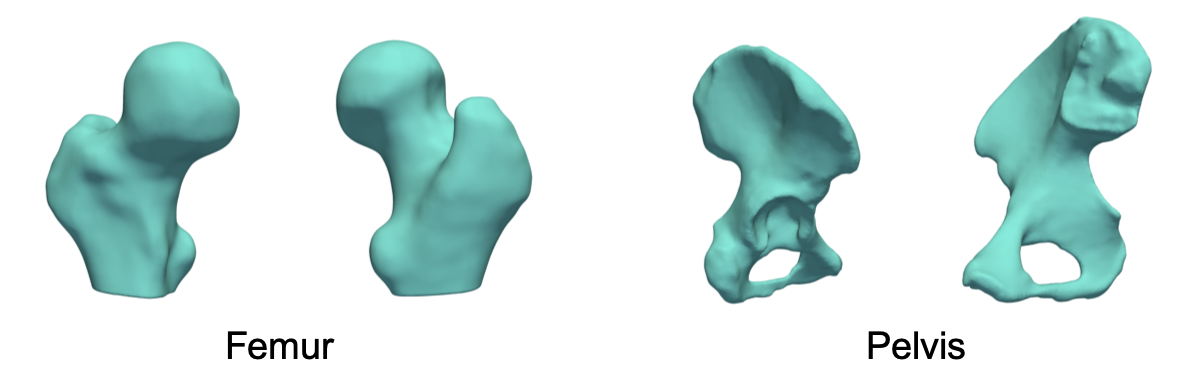}
	\caption{Anatomical complexity in sample femur and pelvis shapes.}
		\label{fig:shapes}
\end{figure}

Using the coffee bean and human scapula datasets, we demonstrated the improved performance of statistical shape models using the proposed deep learned features in the PSM optimization. However, for highly complex medical anatomies, labeled training data are often not available. Therefore, the goal is to capitalize on supervised learning of relatively simpler anatomies and adapt the learned features to complex shapes, without supervision.

Here, we use the femur and pelvis anatomies to demonstrate the feature adaptation via domain adversarial learning. Figure~\ref{fig:shapes} shows sample shapes of the two anatomies that have inherently different topological features. Of the two, femur shapes can be successfully modeled using the position-based shape models \cite{harris2013statistical}, and hence is used as the source anatomy or domain. Pelvis shape, on the other hand, has complex small-sized shape features and hugely varying curvature. Therefore, it is challenging for the position-based models to generate a good shape-correspondence model. In this work, we showcase automated learning of geometric features for the pelvis shape. Unlike Section~\ref{sec:results1}, the goal here is to learn these features without requiring a correspondence model for the pelvis. We first train a CNN in the Siamese framework, using the labeled samples from a femur correspondence model. The trained network is then refined/adapted using the unsupervised domain adversarial training, to adapt to the pelvis anatomy (as the target domain). As a result, the pretrained features are evolved into a combined feature space of femur and pelvis anatomies. 

\begin{figure}[!ht]
	\centering
	\includegraphics[width=0.8\columnwidth]{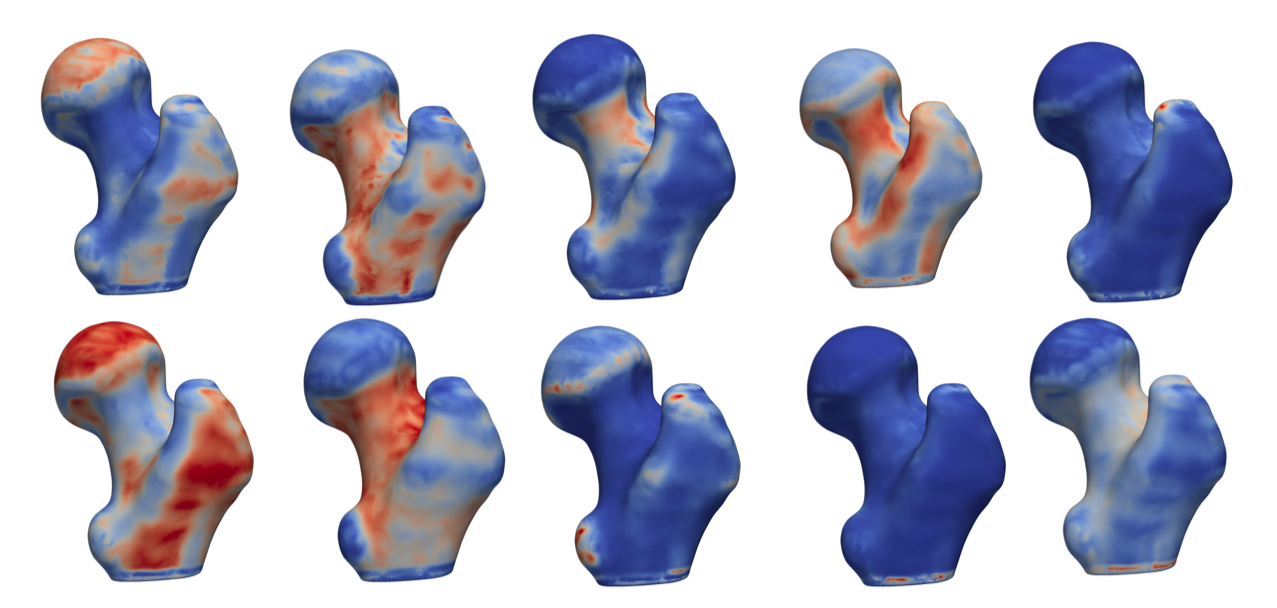}
	\caption{Sample features for the femur shape, learned using the Siamese network.}
		\label{fig:femurFeats}
\end{figure}

We use an optimized statistical shape model with $2045$ correspondence points on a dataset of 8 femur shapes to train the Siamese network. The CNN configuration with $10$ output features performed best for corresponding/noncorresponding classification, yielding a performance of $0.90$ AUC on training samples ($80$\% of the total samples) and $0.87$ AUC on the validation set ($10$\% of the total samples).  Figure~\ref{fig:femurFeats} showcases sample features from the network. The learned features highlight key anatomical aspects of the femur shape, similar to the coffee bean and scapula shapes in Figure~\ref{fig:cnnFeatures}. 

\begin{figure}[!ht]
	\centering
	\includegraphics[width=0.8\columnwidth]{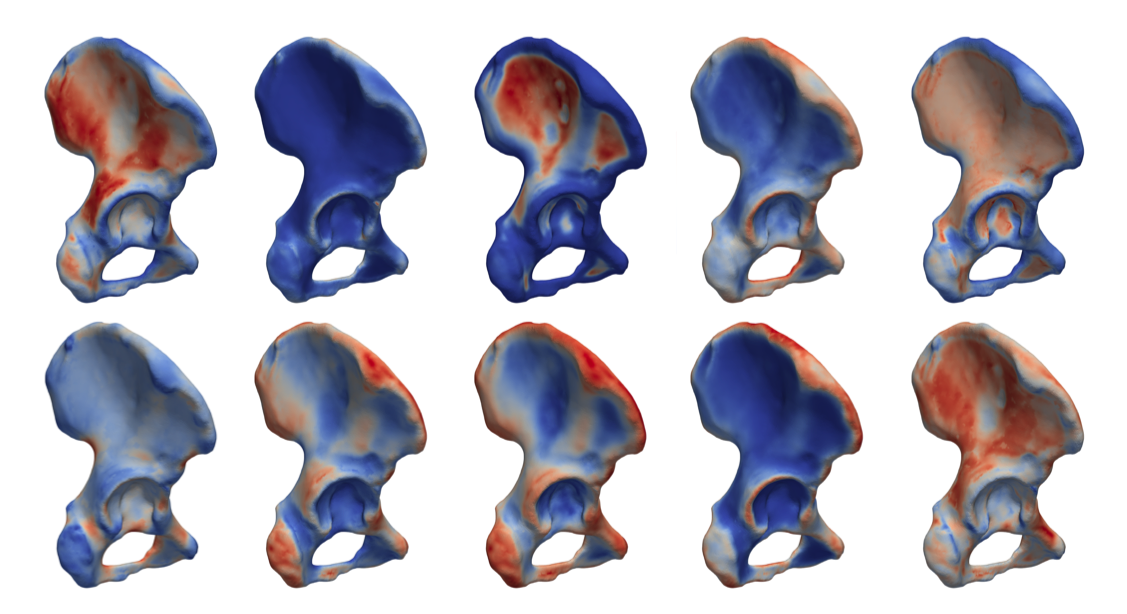}
	\caption{Sample pelvis features obtained via supervised training using the quality-controlled pelvis correspondence model.}
	\label{fig:pelvisOwnFeats}
\end{figure}

\begin{figure}[!ht]
	\centering
	\includegraphics[width=0.8\columnwidth]{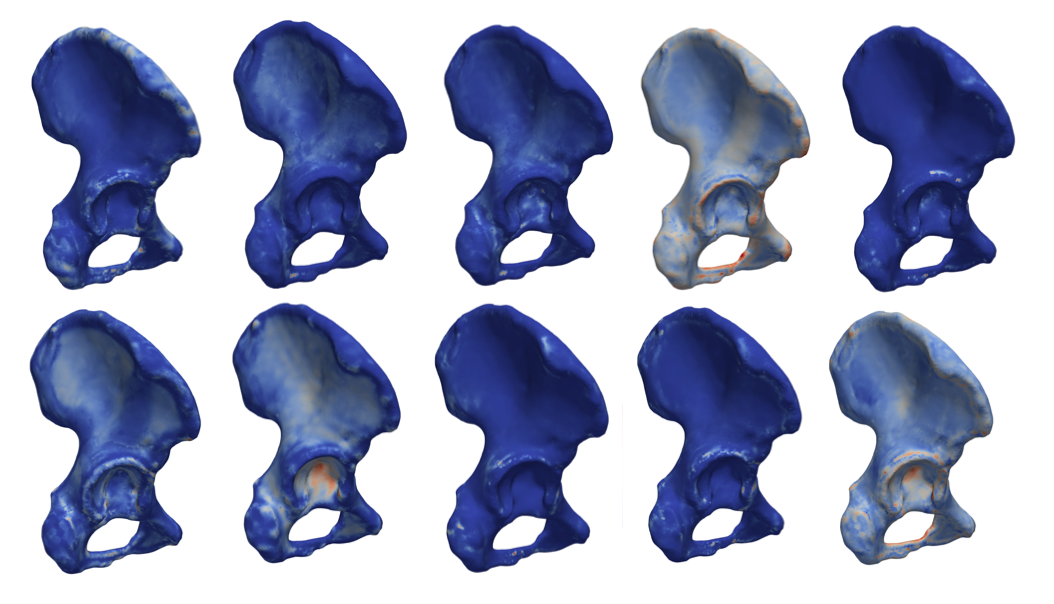}
	\caption{Sample features extracted on the pelvis shape using the CNN trained on the femur shape.}
	\label{fig:femurOnlyPelvisFeats}
\end{figure}

To validate the pelvis features, we independently train a Siamese network using a quality-controlled model of 8 pelvis shapes and $2048$ correspondence points. Figure~\ref{fig:pelvisOwnFeats} shows the sample features thus obtained. To highlight the need for feature adaptation, we use the CNN trained on femur shapes to extract features for the pelvis shape without adaptation; Figure~\ref{fig:femurOnlyPelvisFeats} shows the output feature maps. In comparison, the feature maps extracted using the femur trained model (Figure~\ref{fig:femurOnlyPelvisFeats}) do not generalize for the pelvis anatomy. 

\begin{figure}[!ht]
	\centering
	\includegraphics[width=0.8\columnwidth]{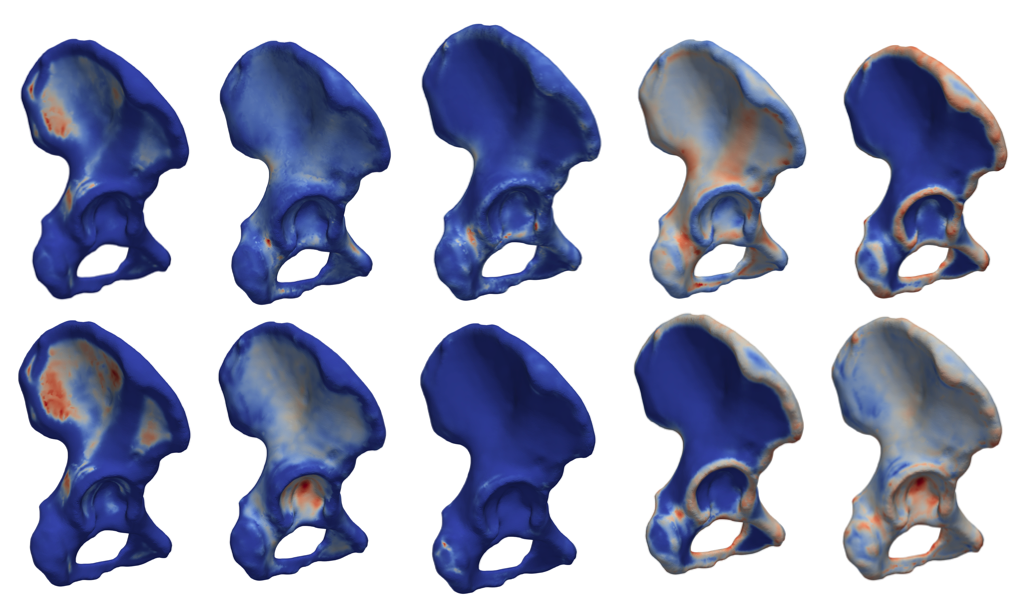}
	\caption{Sample adapted features extracted for the pelvis shape.}
	\label{fig:dannPelvis}
\end{figure}

The femur trained model is then adapted using the domain adversarial training. Figure~\ref{fig:dannPelvis} shows the adapted features extracted on the pelvis shape. As a result of the feature adaptation, the new features are able to model the high curvature regions and the flat anatomical regions in the iliac wing of pelvis. We use the deep learning features to optimize for the shape correspondence model for a dataset of 23 pelvis shapes. The proposed deep learning features are trained using the geodesic patch, which captures the local surface geometry and is invariant to the location and orientation of the shapes. Such invariance is useful for the generalizability of trained networks onto new datasets. As discussed earlier, we use position information with the deep learning features to resolve the correspondence in similar anatomical regions across the shape. For the pelvis anatomy, orientation information plays a vital role in resolving complex regions, such as the very thin iliac wing. Therefore, we use the position and orientation information along with the deep features in PSM optimization. Specifically, we use the following combinations of position, orientation (surface normals), and deep learning features to compare the performance:
\begin{itemize}
	\item XYZ: positional PSM \cite{cates2007shape,cates2017shapeworks};
	
	\item Normals: position and surface normals in the PSM framework;
	
	\item FemurFea: position and femur-trained features (Figure~\ref{fig:femurOnlyPelvisFeats}) in the PSM framework;
	
	\item PelvisFea: position and pelvis-trained features (Figure~\ref{fig:pelvisOwnFeats}) in the PSM framework;
	
	\item DannFea: position and domain-adapted features (Figure~\ref{fig:dannPelvis}) in the PSM framework;
	
	\item FemurFea-normals: position, surface normals, and femur-trained features (Figure~\ref{fig:femurOnlyPelvisFeats}) in the PSM framework;
	
	\item PelvisFea-normals: position, surface normals, and pelvis-trained features (Figure~\ref{fig:pelvisOwnFeats}) in the PSM framework; and
	
	\item DannFea-normals: position, surface normals, and domain-adapted features (Figure~\ref{fig:dannPelvis}) in the PSM framework.
\end{itemize}

\begin{figure}[!ht]
	\centering
	\includegraphics[width=0.8\columnwidth]{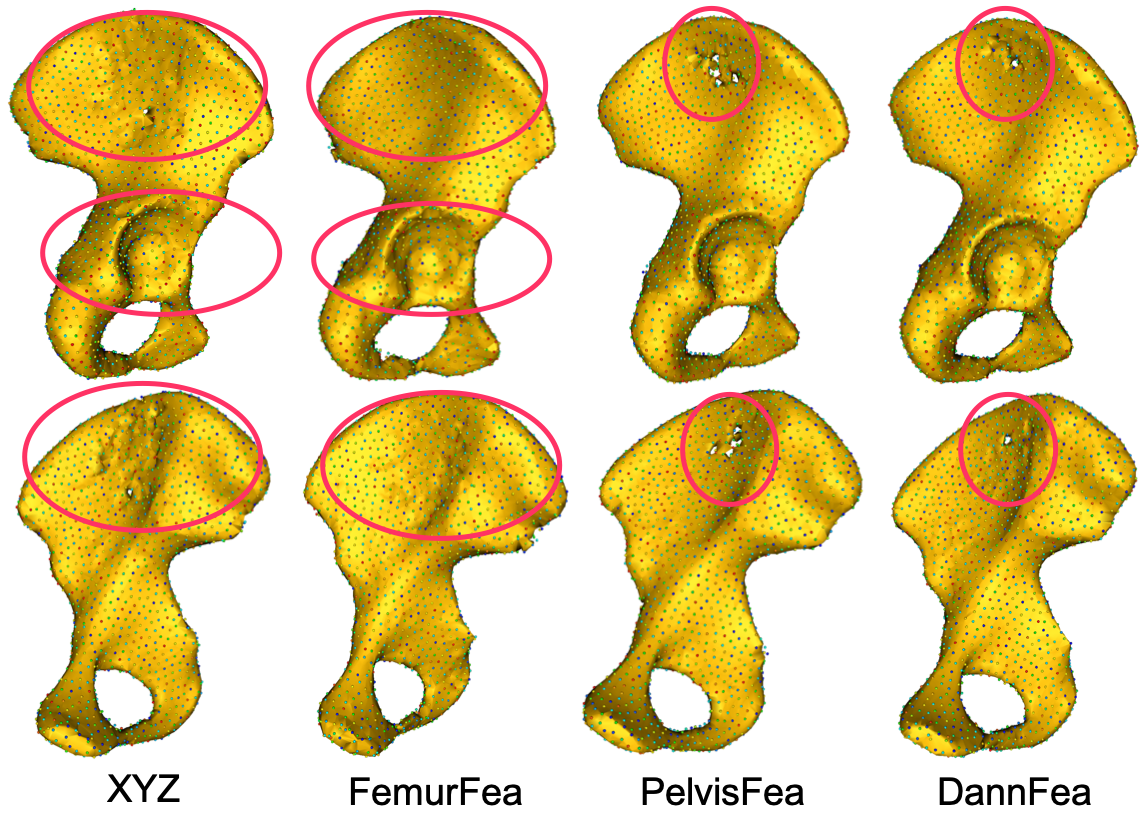}
	\caption{Comparing mean shapes from positional and the three deep feature PSM models. The highlighted regions contain incorrect correspondence points.}
	\label{fig:meanPelvis-positional}
\end{figure}
\begin{figure}[!ht]
	\centering
	\includegraphics[width=0.8\columnwidth]{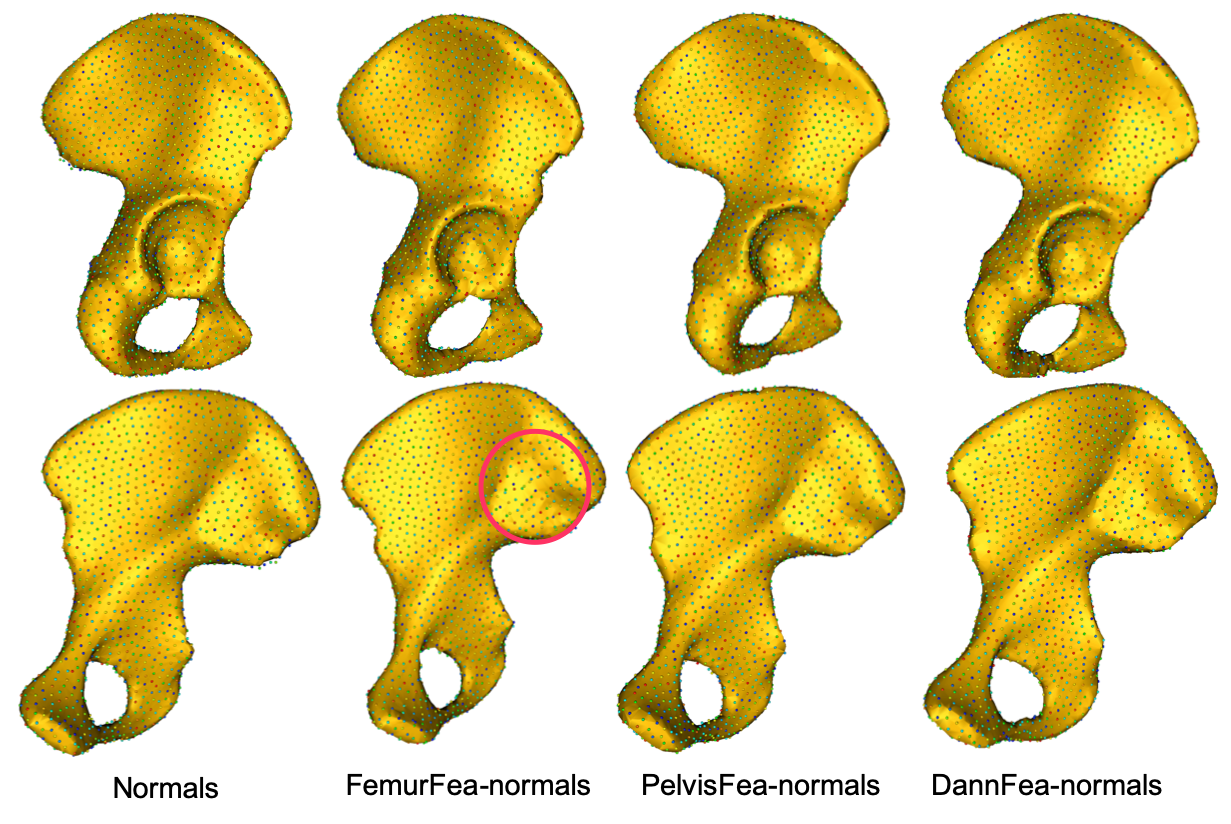}
	\caption{Comparing mean shapes after including normals to positional and deep feature PSM models. The highlighted regions contain incorrect correspondence points.}
	\label{fig:meanPelvis-normals}
\end{figure}

\begin{figure}[!ht]
	\centering
	\includegraphics[width=0.83\columnwidth]{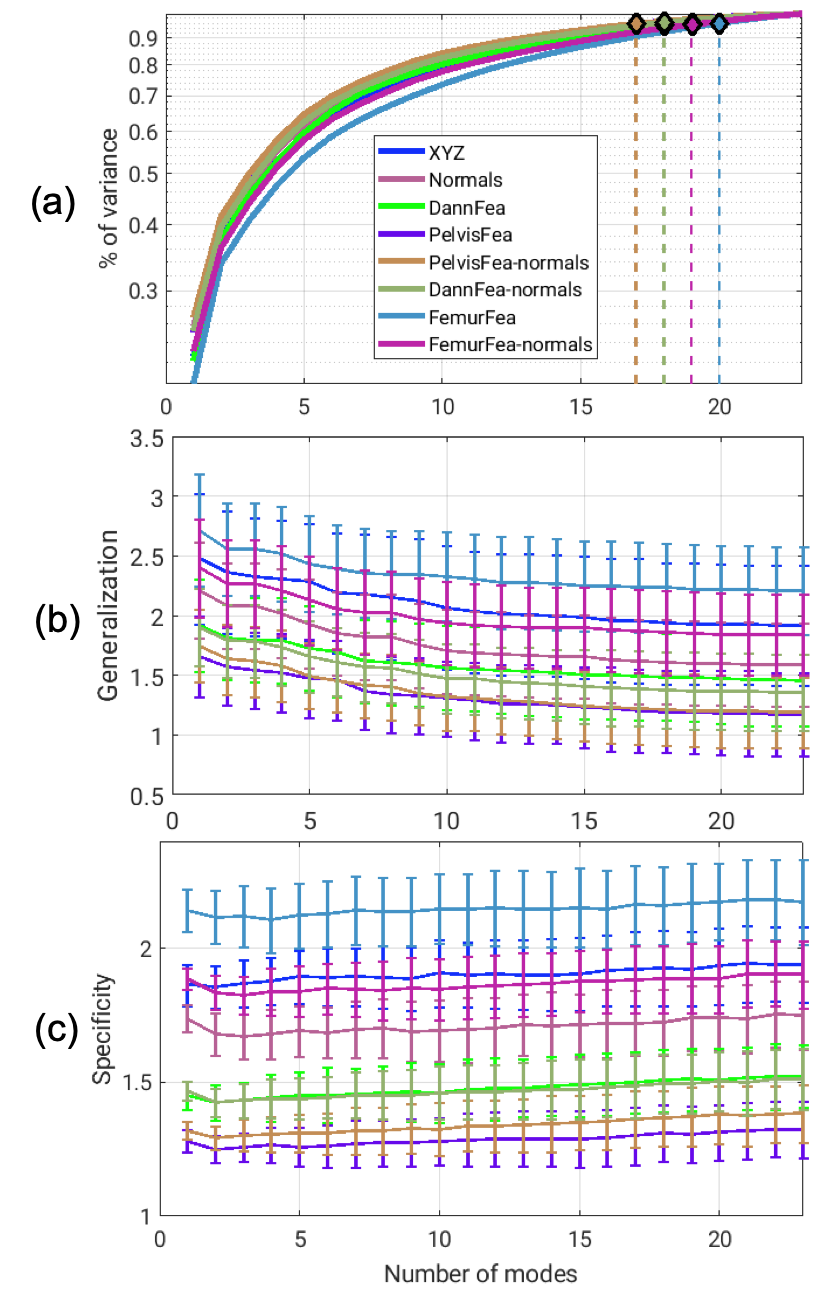}
	\caption{Pelvis results: (a) variance plot, (b) generalization (mm), (c) specificity (mm), voxel size=0.5mm.}
	\label{fig:quantPelvis}
\end{figure}

The pelvis is a complex shape with very thin structures, and therefore, both position and orientation provide crucial support to other geometric features. First, we showcase the mean shapes generated using models involving position and deep learning features; see Figure~\ref{fig:meanPelvis-positional}. As expected, the pelvis-trained and domain-adapted features are able to significantly reduce the number of incorrect correspondences compared to the positional PSM (XYZ) and the FemurFea models. The few unresolved correspondences in the PelvisFea and DannFea models belong to the highly thin region (up to a voxel thickness of 0.5 mm) in the iliac wing (highlighed by the red circle in Figure~\ref{fig:meanPelvis-positional}).  Adding surface normals to the PSM further resolves all the incorrect correspondences; see Figure~\ref{fig:meanPelvis-normals}. Figure~\ref{fig:quantPelvis} presents the quantitative comparison of all PSM variants. Figure~\ref{fig:quantPelvis}(a) compares the variance plots for the different models. PelvisFea-normals produce the most compact model, followed by the PelvisFea, DannFea-normals, normals, DannFea, XYZ, FemurFea-normals, FemurFea, respectively. This trend highlights that the adapted features provide an improvement over traditional PSM models (positional and position with surface normals), and are upper-bounded by the performance of deep features learned with supervision (PelvisFea and PelvisFea-normals). The quantitative performance plots in Figure~\ref{fig:quantPelvis}(b-c) closely follow a similar trend, thus highlighting the need for domain-specific geometric features to generate improved shape correspondence models.
\end{sloppypar}

%

	\section{Conclusion}
\begin{sloppypar}
This paper proposed deep learning methods to resolve the challenges in shape correspondence estimation for complex anatomies. Supervised and unsupervised learning approaches are proposed to facilitate geometric feature learning. The unsupervised technique is especially useful when a training correspondence model is difficult to obtain for complex anatomies. Thus far, such models have been generated with the help of manual expertise, which may limit the throughput of studies with a large number of sample shapes. Quantitative results obtained using synthetic and clinical datasets showcase the success of the proposed method to improve shape correspondence models. The results showed that position and orientation still play a crucial role in the correspondence optimization. Next steps should involve learning geometric features that also encapsulate this global information and thus reduce the high computation cost of Hessian, required for any normals-based PSM. Furthermore, image-based feature learning can help reduce the preprocessing overhead of shape segmentation.
\end{sloppypar}
	
	\bibliographystyle{IEEEtran}
	\bibliography{refs.bib}

\end{document}